\documentclass[sigconf]{acmart}
\usepackage{bm}
\usepackage{amsfonts}
\usepackage{hyperref}
\usepackage{url}
\usepackage{graphicx}
\usepackage{tcolorbox}
\usepackage{xcolor}
\usepackage{tikz}
\usepackage{booktabs}
\usepackage{mdframed}
\usepackage{xcolor}
\usepackage{pifont}
\usepackage{xspace}
\usepackage{multirow}
\usepackage{wrapfig}
\usepackage{bm}
\usepackage{booktabs}       
\usepackage{amsfonts}       
\usepackage{nicefrac}       
\usepackage{microtype}      
\usepackage{xcolor}    
\usepackage{graphicx}
\usepackage{subfigure} 
\usepackage{amsmath}

\usepackage{amssymb}
\usepackage{multicol}
\usepackage{balance}
\usepackage{enumitem}
\usepackage{array}
\usepackage{comment}
\usepackage{caption}
\usepackage{adjustbox}
\usepackage{amsthm}
\usepackage{mwe}
\AtBeginDocument{%
  }

\copyrightyear{2025}

\acmYear{2025}

\setcopyright{rightsretained}

\acmConference[KDD '25] {Proceedings of the 1st Workshop on "AI for Supply Chain: Today and Future" @ 31st ACM SIGKDD Conference on Knowledge Discovery and Data Mining V.2}{August 3, 2025}{Toronto, ON, Canada.}

\acmBooktitle{Proceedings of the 1st Workshop on "AI for Supply Chain: Today and Future" @ 31st ACM SIGKDD Conference on Knowledge Discovery and Data Mining V.2 (KDD '25), August 3, 2025, Toronto, ON, Canada}

\acmISBN{979-8-4007-1454-2/25/08}

\acmDOI{10.1145/XXXXXX.XXXXXX}

\begin{document}

\title{Time Series Forecastability Measures}


\author{Rui Wang}
\affiliation{%
  \institution{Amazon Web Services}
  \city{Seattle}
  \state{WA}
  \country{USA}
  }
\email{rwngamz@amazon.com}

\author{Steven Klee}
\affiliation{%
  \institution{Amazon Web Services}
  \city{Bellevue}
  \state{WA}
  \country{USA}
}
\email{sklee@amazon.com}

\author{Alexis Roos}
\affiliation{%
  \institution{Amazon Web Services}
  \city{Seattle}
  \state{WA}
  \country{USA}
}
\email{alexiroo@amazon.com}





\renewcommand{\shortauthors}{Trovato et al.}

\begin{abstract}
This paper proposes using two metrics to quantify the forecastability of time series prior to model development: the spectral predictability score and the largest Lyapunov exponent. Unlike traditional model evaluation metrics, these measures assess the inherent forecastability characteristics of the data before any forecast attempts. The spectral predictability score evaluates the strength and regularity of frequency components in the time series, whereas the Lyapunov exponents quantify the chaos and stability of the system generating the data. We evaluated the effectiveness of these metrics on both synthetic and real-world time series from the M5 forecast competition dataset. Our results demonstrate that these two metrics can correctly reflect the inherent forecastability of a time series and have a strong correlation with the actual forecast performance of various models. By understanding the inherent forecastability of time series before model training, practitioners can focus their planning efforts on products and supply chain levels that are more forecastable, while setting appropriate expectations or seeking alternative strategies for products with limited forecastability.
\end{abstract}

\maketitle

\section{Introduction}
In the rapidly evolving landscape of supply chain management, accurate time series forecasting has become an indispensable tool for demand prediction, inventory optimization, and supply planning \cite{lim2021time, liang2024foundation, rangapuram2018deep, benidis2022deep, hamilton2020time}. However, the effectiveness of these forecasts is intrinsically tied to the inherent forecastability of the underlying data. Not all time series exhibit the same degree of forecastability, and this variability can significantly impact the reliability of business decisions based on these predictions.

Traditionally, practitioners assess forecastability post hoc—by training models and evaluating performance. Although effective, this process is computationally expensive and can lead to wasted effort in inherently unpredictable series. We propose a more systematic alternative: using spectral predictability \cite{goerg2013forecastable} and Lyapunov exponents \cite{dingwell2006lyapunov} to quantify a time series’ forecastability a priori \cite{wang2023koopman}. We will demonstrate how these metrics can be systematically applied to time series data to identify the inherent difficulty of forecasting tasks and support better planning and resource allocation.

The spectral predictability evaluates the strength and complexity of frequency components within a time series, providing insights into its underlying patterns and cyclicality. Lyapunov exponent analysis, on the other hand, measures the stability and chaos of the data-generating system, offering insight into long-term behavior. Together, they offer complementary views into a series’ structure and long-term dynamics.

This approach is particularly useful in supply chain management \cite{aviv2003time, mentzer2001defining, power2005supply}, where data is highly heterogeneous between products, categories, and regions. By understanding the forecastability of time series at various aggregation levels—such as individual products, product categories, or regional sales—decision-makers can better navigate complex networks, focus modeling efforts on more predictable areas, allocate resources efficiently, and set realistic expectations for forecasting performance.

We validate the use of these metrics through experiments on both synthetic and real-world datasets. In synthetic data, we show that spectral predictability and Lyapunov exponents strongly correlate with the underlying complexity of time series, effectively distinguishing between simple, noisy, chaotic, and random signals. In the hierarchical M5 dataset, we observed strong correlations between forecastability scores and actual forecast performance at different aggregation levels. Together, these findings demonstrate that the proposed use of these metrics offers a practical and computationally efficient way to assess time-series forecastability and guide forecasting strategies. They can set expectations on forecast performance and potentially inform hedging or intervention strategies, such as using different models for items with different levels of forecastability. Furthermore, these metrics provide valuable insights into model performance, offering a theoretical framework to explain why certain predictive models succeed or fail across different types of time series.

\section{Methodology}
We describe two metrics—Spectral Predictability and the largest Lyapunov Exponent—used to assess a time series' forecastability prior to model training. We provide detailed explanations of how each metric is computed and interpreted in the context of identifying intrinsic predictability in time series data.

\subsection{Spectral Predictability}
Spectral Predictability \cite{goerg2013forecastable} quantifies the concentration and regularity of frequency components in a time series, serving as a proxy for its complexity in the Fourier domain. Time series with clear periodic patterns (e.g., seasonality) exhibit dominant frequency peaks, while highly irregular or noisy series have energy dispersed across a wide range of frequencies. In this context, predictability is inversely related to the spectral entropy—a measure of disorder in the frequency domain.

Given a de-trended time series, $\bm{y} = (y_0, y_1, \dots ,y_{T-1})$, we first compute its power spectral density (PSD) using the Fast Fourier Transform \cite{duhamel1990fast}. Let $p_i$ denote the normalized power of the $i$-th frequency component. The spectral entropy is given by:
\begin{equation}
H_a(\bm{y}) = \sum_i p_i \log_a p_i
\end{equation}

The Spectral Predictability score can be defined as:
\begin{equation}
\Omega(\bm{y}) = 1 - \frac{H_a(\bm{y})}{\log_a(2\pi)},
\end{equation}

where $a$ is the logarithmic base, typically set to $e$ or 2.
Normalizing by $\log_a(2\pi)$ bounds $\Omega(\bm{y})$ in $[0,1]$, with higher values indicating lower spectral complexity and greater forecastability

The intuition behind this metric is that the complexity of a time series in the Fourier domain is directly related to its forecastability. For example, a flat spectrum indicates high unpredictability, as maximum spectral entropy corresponds to a uniform distribution of energy across all frequencies, where all possible frequencies contribute equally to the time series, making it highly complex and difficult for any model to forecast. Conversely, a constant time series exhibits zero spectral entropy and therefore has the highest spectral predictability.

To mitigate spectral leakage, we apply a Hann window before computing the Fourier transform \cite{pielawski2020introducing, lyon2009discrete}. The metric can be computed globally or within a moving window to detect local changes in predictability. It is computationally efficient with time complexity $O(T\log T)$, making it practical for large-scale analysis.

\subsection{Lyapunov Exponents}
While spectral predictability captures harmonic structure in the frequency domain, it does not differentiate between deterministic chaos and stochastic noise. To address this, we complement it with Lyapunov Exponents \cite{dingwell2006lyapunov}, which measure the sensitivity of a dynamical system to initial conditions in the time domain. This metric provides insight into the system’s stability and long-term behavior.

Given a time series $\bm{y} = (y_0, y_1, ... ,y_{T-1})$, we first reconstruct its state space via time-delay embedding:
\begin{equation}
    \mathbf{x}_t = \left( y_t, y_{t+\tau}, \dots, y_{t+(m-1)\tau} \right) \in \mathbb{R}^m
\end{equation}
where $m$ is the embedding dimension and $\tau$ is the delay. Each vector $\mathbf{x}_t$ represents the system’s state at time $t$ in the reconstructed phase space.

To estimate the largest Lyapunov exponent, we track how the distance between initially close state vectors diverges over time.  For each embedded state $\mathbf{x}_t$, we identify its nearest neighbor $\mathbf{x}_t'$, with initial separation:
\begin{equation}
\delta_0 = \| \mathbf{x}_t - \mathbf{x}_t' \|
\end{equation}

We then observe how this separation evolves over a fixed number of time steps $\Delta t$:
\begin{equation}
\delta(\Delta t) = \| \mathbf{x}_{t+\Delta t} - \mathbf{x}_{t'+\Delta t} \|
\end{equation}

The largest Lyapunov exponent is estimated as the average exponential rate of divergence:

\begin{equation}
\lambda = \frac{1}{\Delta t} \log \frac{\|\delta(\Delta t)\|}{\|\delta_0\|}
\end{equation}

A positive $\lambda$ indicates exponential divergence and chaotic behavior, implying reduced forecastability. A non-positive $\lambda$ (zero or negative) suggests stability and higher forecastability.

In practice, we average $\lambda$ over multiple state pairs to improve robustness. Since the estimation depends on accurate local trajectory tracking, it requires a sufficiently long and dense time series. Based on our experiments, we recommend using at least $100 \times m$ data points and limiting sparsity to below 0.7. This method is more computationally intensive than spectral analysis, typically with $O(T^2)$ complexity, but provides valuable insight into chaotic behaviors of a time series

Note that a dynamical system has a full spectrum of Lyapunov exponents 
\(\boldsymbol{\lambda} = (\lambda_1, \lambda_2, \dots, \lambda_m)\), 
one per dimension in the reconstructed phase space. Each exponent 
\(\lambda_i\) measures the average exponential rate of divergence along a specific direction. In this work, we estimate 
only the \emph{largest} Lyapunov exponent, defined as 
\(\lambda = \max_i \lambda_i\), which dominates the system's 
long-term behavior.

\section{Experiments}
We conduct a series of experiments on both synthetic and real-world datasets to evaluate whether spectral predictability and Lyapunov exponents effectively reflect the intrinsic forecastability of time series. Our goals are twofold: (1) to validate that these metrics correlate with time series complexity and predictability, and (2) to demonstrate their alignment with downstream forecasting performance across different levels of data granularity.

\subsection{Forecastability of a Synthetic Example} 
\begin{figure}[htb!]
    \centering
    \includegraphics[width=\linewidth]{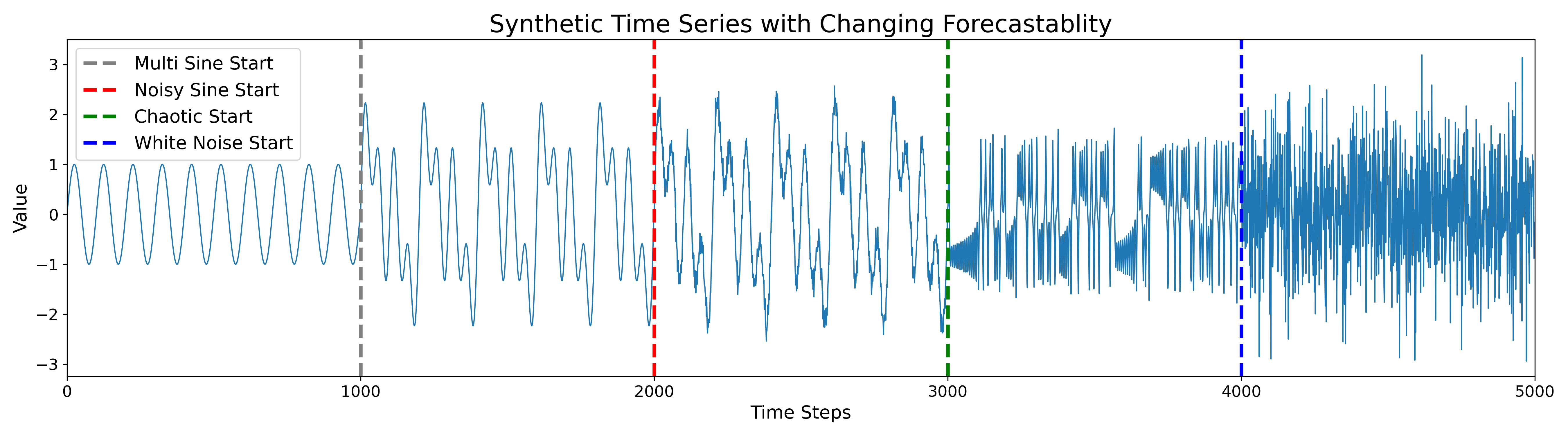}
    \caption{Five segments with increasing complexity and decreasing forecastability: a pure sine wave, a multi-frequency wave, a multi-frequency wave with additional random noise, a trajectory from Lorenz chaotic system and white noise.
}
    \label{fig:synthetic_ts}
\end{figure}

\paragraph{Experiment Setup}
To illustrate the behavior of the two metrics, we construct a synthetic time series composed of five consecutive segments with increasing complexity and decreasing forecastability. These five segments of time series are shown in Figure \ref{fig:synthetic_ts}, including a pure sine wave, a multi-frequency wave, a noisy multi-frequency wave, a Lorenz system trajectory, and white noise.

\begin{figure}[htb!]
    \centering
    \includegraphics[width=\linewidth]{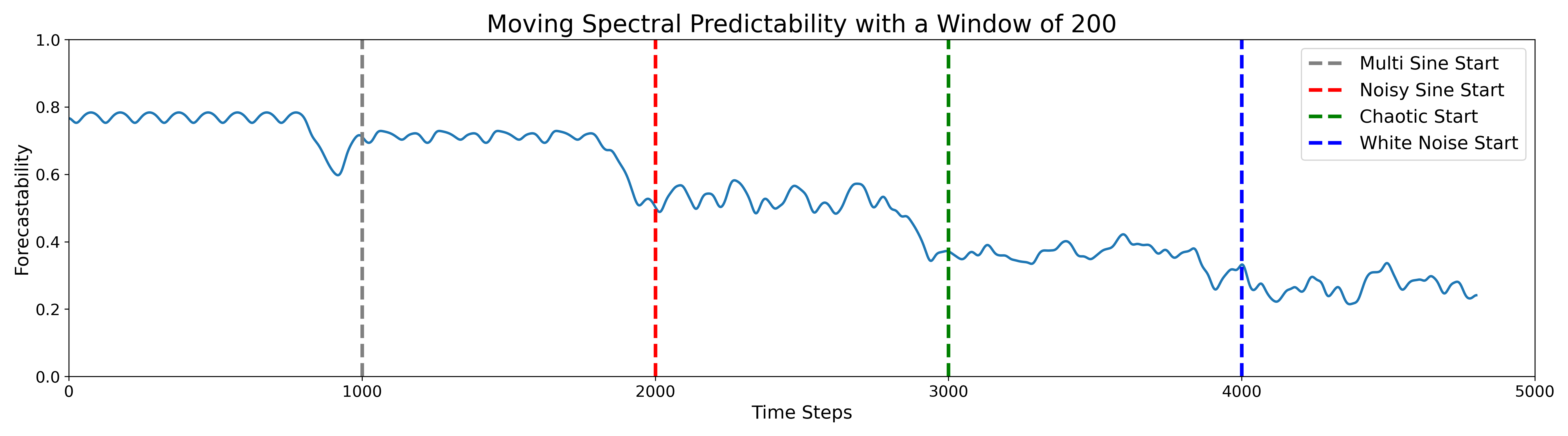}
    \includegraphics[width=\linewidth]{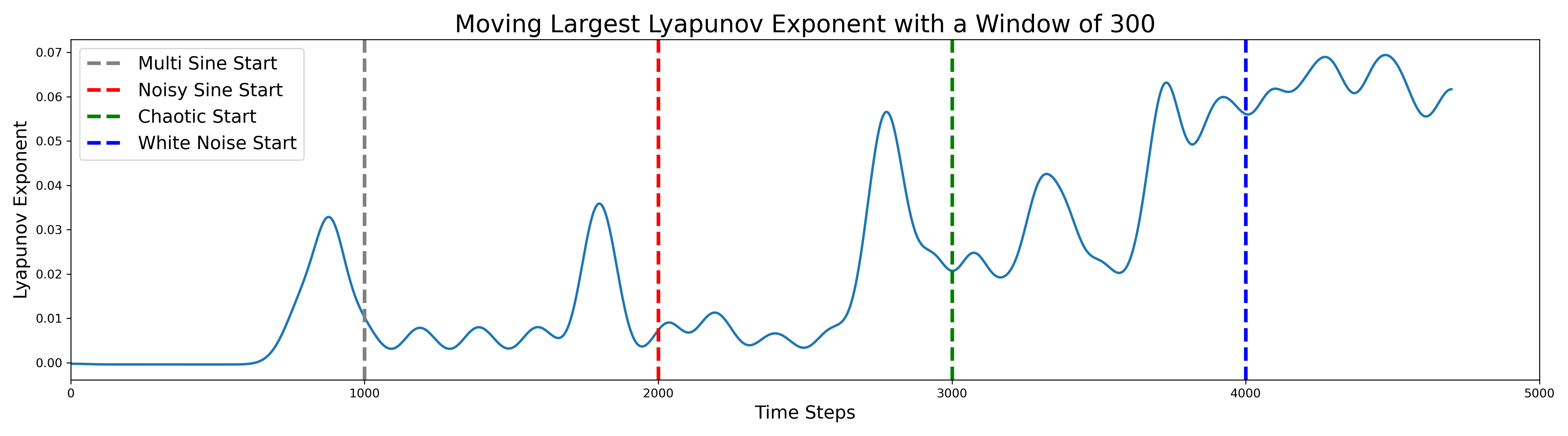}
    \caption{Top: moving spectral predictability with a window size of 200 over the synthetic time series. Bottom: moving largest Lyapunov Exponent with a window size of 300.
}
    \label{fig:moving_forecastability}
\end{figure}

\paragraph{Results} Figure \ref{fig:moving_forecastability} shows the moving spectral predictability with a window size of 200 (top) and moving largest Lyapunov Exponent with a window size of 300 (bottom) computed over the synthetic time series. We can see that both metrics respond consistently with our expectations: spectral predictability decreases, and the Lyapunov exponent increases, as the underlying signal becomes more chaotic or noisy. 

Spectral predictability fluctuates due to the use of a moving window, where each window may not contain full periodic cycles. However, the Fourier transform assumes that the input time series contains complete cycles. Incomplete cycles can introduce noisy spikes in the spectrum. In addition, the sudden increase in the Lyapunov Exponent plot or the sudden drop in Spectral Predictability before each segment occurs because they are computed in a moving window manner. When the window contains two different types of time series, it becomes much harder to forecast.

Note that the reason why the spectral predictability is not zero for the last white noise segment is that it reaches zero only when the spectrum follows a perfectly uniform distribution. This occurs only when the white noise time series is sufficiently long.

While neither metric can distinguish chaos from randomness directly, both serve as strong indicators of overall signal complexity and forecastability. This experiment also highlights their potential utility in identifying distributional shifts or regime changes in non-stationary time series.

\subsection{Sensitivity Study of Metrics to Time Series Length and Sparsity} \label{sec:sensitivity_study}
In this section, we evaluate how the two metrics respond to variations in time series length and sparsity—two key factors often encountered in real-world applications such as retail demand forecasting. Ideally, forecastability scores should decrease as sparsity increases, reflecting the loss of informative structure. We also examine sensitivity to time series length to ensure that, given sufficiently long sequences with similar characteristics, the metrics produce stable and consistent values across different lengths. This stability is important for enabling fair comparisons across different lengths.

\begin{figure}[htb!]
    \centering
    \includegraphics[width=\linewidth]{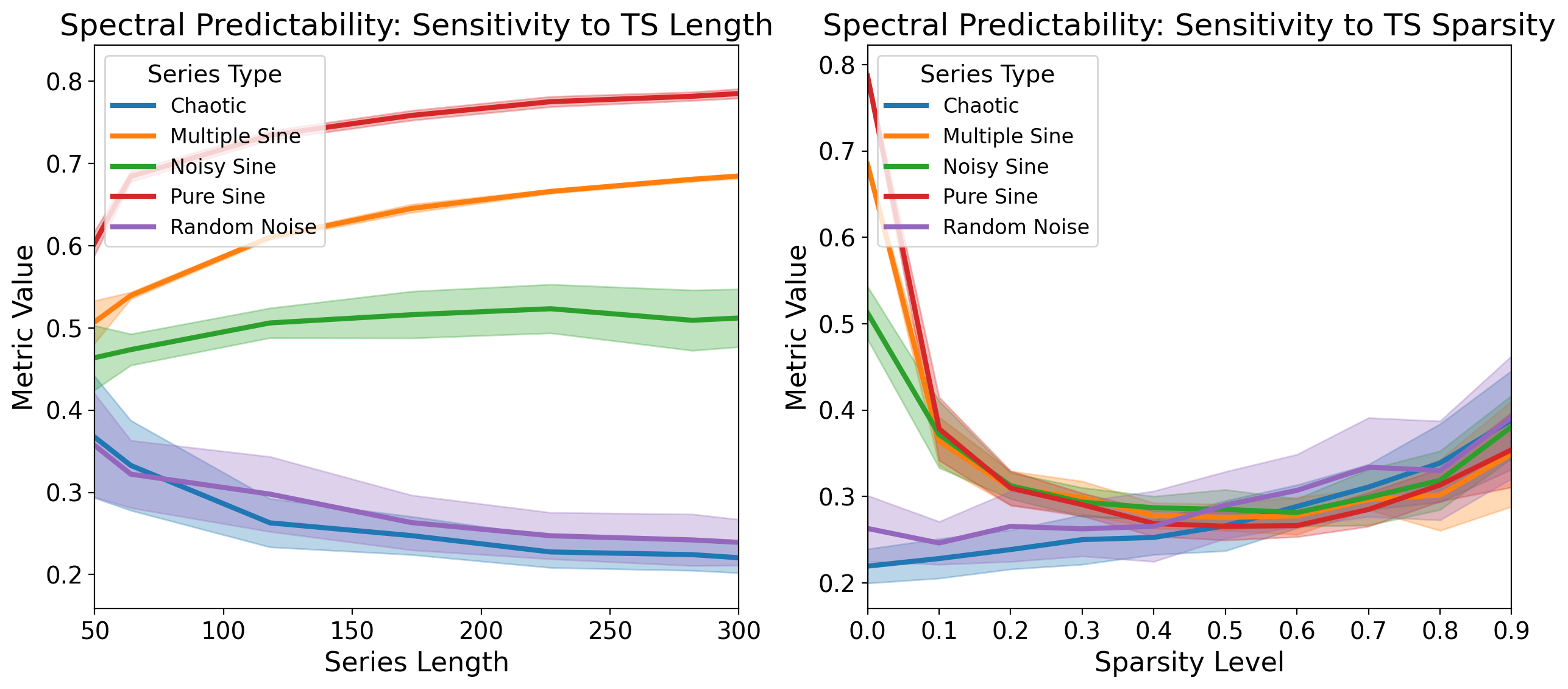}
    \caption{The sensitivity of the spectral predictability to varying time series length and sparsity. (Note: Higher spectral predictability indicates easier-to-forecast series.)
}
    \label{fig:sensitiviy_se}
\end{figure}

\begin{figure}[htb!]
    \centering
    \includegraphics[width=\linewidth]{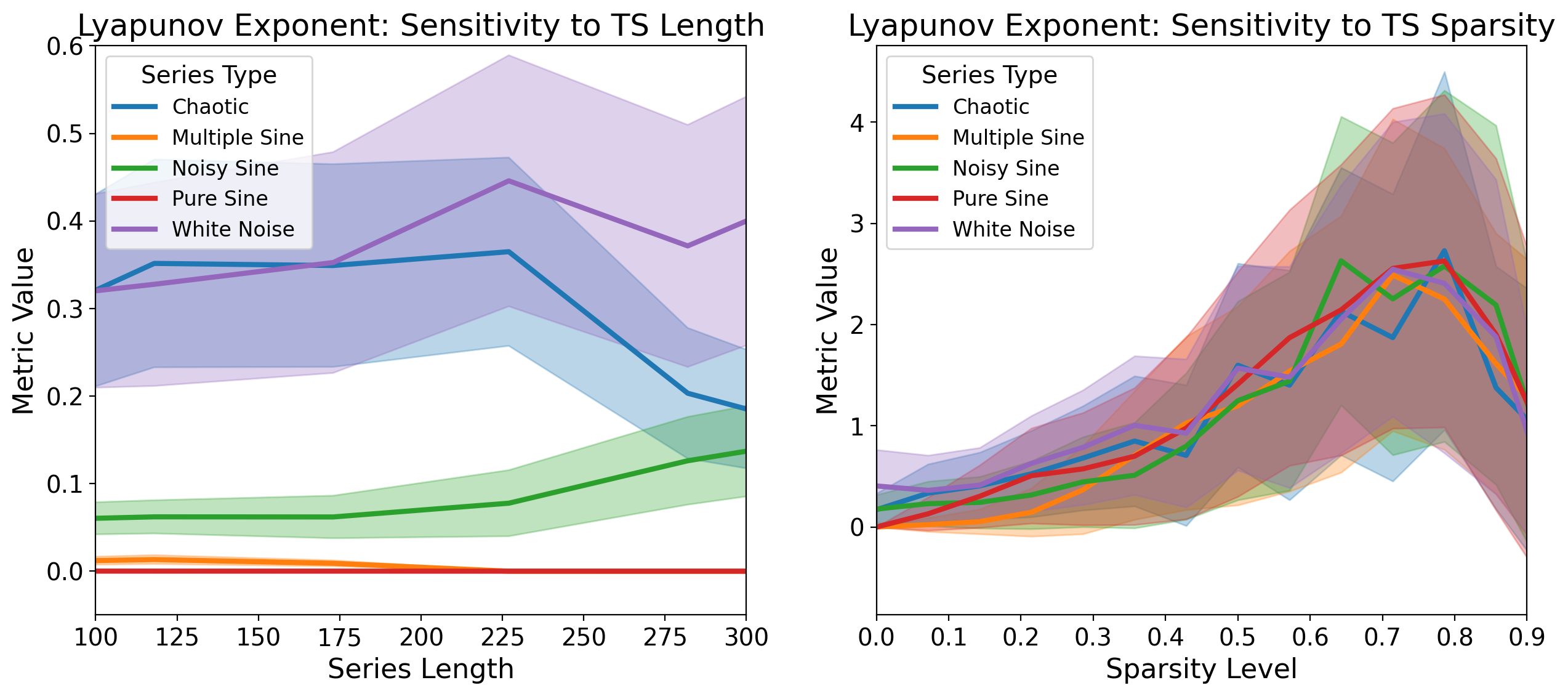}
    \caption{The sensitivity of the Lyapunov Exponent to varying time series length and sparsity. (Reminder: the lower the Lyapunov Exponent, the easier the time series is to forecast)
}
    \label{fig:sensitiviy_le}
\end{figure}

\paragraph{Experiment Setup} We use the same five types of synthetic time series from the previous section, but vary their lengths (from 50 to 300) and sparsity rates (from 0\% to 95\%) by randomly zeroing out values. For each type of synthetic time series, we generate 100 sequences with different initial conditions and system parameters, varying both length and sparsity rate. We then compute spectral predictability and the largest Lyapunov exponent for each configuration. Figure \ref{fig:sensitiviy_se} and Figure \ref{fig:sensitiviy_le}  shows how spectral predictability and Lyapunov Exponent change with series length (left) and sparsity rate (right). The shaded areas in the plots represent two standard deviations.

\paragraph{Results} 

\begin{table*}[htb!]
\centering
\begin{tabular}{|c|c|c||c|c|c|}
\hline
\centering
\textbf{Spectral Predictability} & \textbf{Daily} & \textbf{Weekly} & \textbf{Lyapunov Exponents} & \textbf{Daily} & \textbf{Weekly} \\
\hline
\href{#}{L0 (total)}      & 0.374 $\pm$ 0.0   & 0.394 $\pm$ 0.0   & \href{#}{L0 (total)}      & 0.0 $\pm$ 0.0     & 0.081 $\pm$ 0.0     \\
\href{#}{L1 (category)}   & 0.358 $\pm$ 0.015 & 0.341 $\pm$ 0.008 & \href{#}{L1 (category)}   & 0.0 $\pm$ 0.0     & 0.055 $\pm$ 0.078   \\
\href{#}{L2 (department)} & 0.339 $\pm$ 0.026 & 0.333 $\pm$ 0.047 & \href{#}{L2 (department)} & 0.051 $\pm$ 0.125 & 0.169 $\pm$ 0.112   \\
\href{#}{L3 (product)}    & 0.246 $\pm$ 0.08  & 0.264 $\pm$ 0.057 & \href{#}{L3 (Item)}       & 0.833 $\pm$ 1.495 & 0.231 $\pm$ 0.982   \\
\hline
\end{tabular}
\caption{Spectral Predictability and Lyapunov Exponents across hierarchy levels}
\label{tab:m5}
\end{table*}

We observe several patterns from Figure \ref{fig:sensitiviy_se}. For unpredictable series, longer sequences slightly reduce spectral predictability, while sparsity has a mild inflating effect. For moderately predictable series, predictability remains stable across lengths. For highly predictable series, longer lengths improve predictability, but increased sparsity sharply reduces it. We can conclude that spectral predictability takes sparsity into account and is not significantly affected by length, making it a comprehensive metric for determining forecastability

From Figure \ref{fig:sensitiviy_le}, we can observe that given sufficient length, the Lyapunov exponent isn't affected much by length. Additionally, increasing sparsity will increase the Lyapunov exponent, which is expected as the sparsity make time series harder to predict. But we can also see excessive sparsity (> 0.8) may decreases the Lyapunov exponent and falsely indicates the system as stable. With sufficient length and moderate sparsity, the Lyapunov exponent reliably captures forecastability. However, at extreme sparsity levels (>0.8), it may falsely indicate stability.

\subsection{Forecastability vs. Prediction Errors on the M5 Dataset}
To validate the practical utility of the two metrics, we study their correlation with actual prediction errors on a real-world dataset. We use the M5 forecasting competition dataset, which includes hierarchical sales time series across multiple levels—total sales, state, category, department, and product—at daily granularity.

\paragraph{Experiment Setup}
We compute spectral predictability and Lyapunov exponents for each time series in the M5 dataset across multiple aggregation levels, including total, category, department, and product, as well as two temporal frequencies: daily and weekly. To assess their relationship with actual forecast accuracy, we train several forecasting models—including ETS \cite{gardner1985exponential}, RecursiveTabular \cite{shchur2023autogluon}, and Chronos \cite{ansari2024chronos}—on each subset and compute the Weighted Absolute Percentage Error with help of AutoGluon library \cite{shchur2023autogluon}.

\paragraph{Results} Table \ref{tab:m5} displays the Spectral Predictability and Lyapunov Exponents across multiple hierarchy levels and two temporal frequencies: daily and weekly. We compute the metrics for each time series at each level, and the reported standard deviations reflect the variance across different series within that level. For Spectral Predictability, higher values indicate greater predictability and thus higher forecastability. For the Lyapunov Exponent, a value of zero indicates a stable system, while positive values suggest chaotic behavior and lower forecastability. From the table, we observe that for daily time series, Level 0 (total daily unit sales) exhibits the highest forecastability. Forecastability generally decreases as we move to lower levels in the hierarchy, for both daily and weekly time series. Additionally, for Level 3 (product-level) series, aggregating the data from daily to weekly significantly improves forecastability, as reflected by both metrics.

\begin{figure}[htb!]
    \centering
    \includegraphics[width=\linewidth]{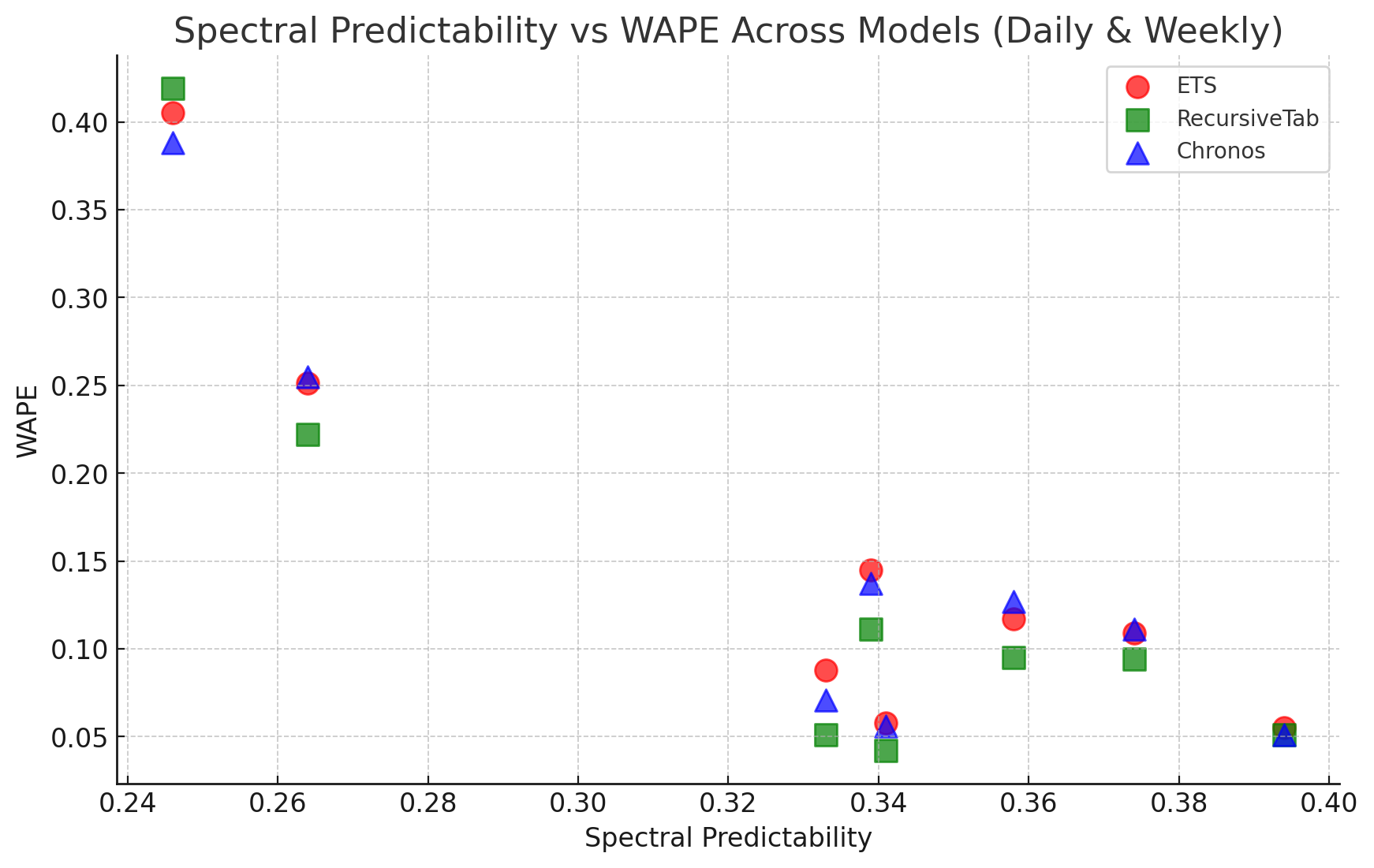}
    \caption{Spectral predictability vs prediction WAPE across multiple levels and time frequencies in M5}
    \label{fig:correlation}
\end{figure}

To investigate whether these observations correlate with actual model prediction errors, we trained multiple models—including ETS, RecursiveTab, and Chronos—on various subsets of the M5 dataset across different hierarchy levels and temporal frequencies. We then plotted their prediction errors against the pre-computed spectral predictability scores, as shown in Figure \ref{fig:correlation}.

Based on Figure \ref{fig:correlation} and Table \ref{tab:m5}, we observe that spectral predictability generally shows a negative correlation with WAPE, suggesting that higher predictability is associated with better model performance. In contrast, Lyapunov exponents exhibit a positive correlation with WAPE, indicating that more chaotic time series are more difficult to forecast. These correlations are strong for both daily and weekly frequencies (with $r = \sim 0.9$).

This experiment, together with the study on the synthetic time series dataset, demonstrates that both forecastability metrics reliably capture the inherent predictability of time series and show strong correlation with downstream forecasting performance. However, we note that this analysis does not provide sufficient evidence to determine which forecasting model performs best under varying levels of forecastability.

\section{Discussion}
We propose using two metrics—Spectral Predictability and the largest Lyapunov Exponent—to assess the forecastability of time series prior to model training. Our goal is to provide a lightweight, model-agnostic way to evaluate whether a time series is inherently predictable, enabling practitioners to focus resources on tractable forecasting tasks and adopt alternative strategies for more chaotic or noisy series.

Our experiments demonstrate that both metrics capture key aspects of time series complexity and correlate strongly with downstream forecasting performance. Spectral Predictability, based on frequency-domain entropy, offers a robust and computationally efficient measure that performs well across a wide range of data conditions. It is particularly useful when time series are short or moderately sparse. Lyapunov Exponents, which quantify sensitivity to initial conditions in the reconstructed phase space, provide complementary insights into the stability of dynamical behavior, but require longer sequences and lower sparsity to be reliable.

As a practical guideline, we find that spectral predictability scores below 0.2 or Lyapunov exponents above 1.0 are indicative of low forecastability. Spectral Predictability is stable with as few as 100 time steps, whereas Lyapunov estimation typically requires at least 100×$m$ time steps, where 
$m$ is the embedding dimension. Comparing spectral predictability to that of white noise with matched length and sparsity further improves interpretability. Practitioners should exercise caution when interpreting results for highly sparse or short time series, such as monthly or yearly retail data, where metric stability may degrade.

Beyond guiding modeling strategy, these metrics have broader utility. Persistently low or unstable values can signal data quality issues such as insufficient history or structural noise. When computed over sliding windows, they can serve as indicators of distributional shift or regime change, helping determine when to retrain forecasting models. In this way, they also support model interpretability and monitoring.

Future work includes integrating these metrics into automated model selection pipelines and using them to inform active learning, anomaly detection, and dynamic retraining in real-time forecasting systems.
\newpage
\bibliographystyle{ACM-Reference-Format}
\bibliography{sample-base.bib}
\end{document}